# UzbekTagger: The rule-based POS tagger for Uzbek language

Maksud Sharipov[1], Elmurod Kuriyozov[1,2], Ollabergan Yuldashev[1], Og'abek Sobirov[1]

[1] Urgench State University, Department of Information Technologies; 14, Kh.Alimdjan str, Urgench, Uzbekistan;
{m.sharipov,elmurod1202,ollaberganyuldashov,sobirov_o}@urdu.uz

[2] Universidade da Coruña, CITIC; Campus de Elviña, A Coruña, 15071, Spain;
e.kuriyozov@udc.es

**Abstract**

This research paper presents a part-of-speech (POS) annotated dataset and tagger tool for the low-resource Uzbek language. The dataset includes 12 tags, which were used to develop a rule-based POS-tagger tool. The corpus text used in the annotation process was made sure to be balanced over 20 different fields in order to ensure its representativeness. Uzbek being an agglutinative language so the most of the words in an Uzbek sentence are formed by adding suffixes. This nature of it makes the POS-tagging task difficult to find the stems of words and the right part-of-speech they belong to. The methodology proposed in this research is the stemming of the words with an affix/suffix stripping approach including database of the stem forms of the words in the Uzbek language. The tagger tool was tested on the annotated dataset and showed high accuracy in identifying and tagging parts of speech in Uzbek text. This newly presented dataset and tagger tool can be used for a variety of natural language processing tasks such as language modeling, machine translation, and text-to-speech synthesis. The presented dataset is the first of its kind to be made publicly available for Uzbek, and the POS-tagger tool created can also be used as a pivot to use as a base for other closely-related Turkic languages.

## 1. Introduction

Part-of-Speech (POS) tagging is a process of identifying and labeling the grammatical category of each word in a given text. POS tagging is a fundamental task in natural language processing (NLP) and is used in a wide range of applications such as text analysis, machine translation, language modeling, and information retrieval. It is also a key step in many other NLP tasks, such as syntactic parsing, named entity recognition, and sentiment analysis.

POS tagging has evolved from rule-based systems (Kupiec, 1992) to machine learning-based models (Awasthi et al., 2006; Constant et al., 2011), and now deep learning-based models (dos Santos et al., 2014; Meftah et al., 2018; Perez-Ortiz et al., 2001). It is widely used in various applications and continues to be an active area of research in the field of NLP (Manning, 2011).

In this research, we present UzbekTagger - a Part-of-Speech (POS) tagger tool and an annotated dataset for the Uzbek language. Firstly, we carefully analysed the previous studies as well as the linguistic nature of the language under focus, and decided 12 POS tags to be used. Then, the rule-based POS-tagger tool for Uzbek, called UzbekTagger, was created in Python. The tool is based on stems and suffix-affix data and rules in our codebase, allowing for efficient and accurate tagging of given text in Uzbek. Lastly, we manually annotated a special Uzbek corpus which was balanced over 23 distinct fields with ~1K words each to ensure its representative nature. The POS-tagging guidelines of Universal Dependencies (version 2)[1] were followed during the creation of the tagger.

The POS tags proposed for Uzbek in this work are twelve main categories are following:
- **Open word classes:** noun, verb, adjective, numeral, adverb, pronoun;
- **Closed word classes:** auxiliary, conjunction, particle;
- **Intermediate words:** modal words, imitation words, interjection words.

All the identifier names of the proposed tags, their meaning and example Uzbek words are given in Table 1.

| Id | Tag | Meaning | Uzbek examples |
|---|---|---|---|
| 1 | NOUN | OT (*Noun*) | olma (*apple*) |
| 2 | VERB | FE'L (*Verb*) | yugurmoq (*run*) |
| 3 | ADJ | SIFAT (*Adjective*) | ko'p (*many/much*) |
| 4 | NUM | SON (*Numeral*) | besh (*five*) |
| 5 | ADV | RAVISH (*Adverb*) | tez (*fast*) |
| 6 | PRON | OLMOSH (*Pronoun*) | bu (*this*) |
| 7 | AUX | KO'MAKCHI (*Auxiliary*) | bilan (*with*) |
| 8 | CONJ | BOG'LOVCHI (*Conjunction*) | va (*and*) |
| 9 | PART | YUKLAMA (*Particle*) | faqat (*only*) |
| 10 | MOD | MODAL (*Modal*) | darhaqiqat (*actually*) |
| 11 | IMIT | TAQLID (*Imitation*) | kuk-kuk (*imitation of a hen*) |
| 12 | INTJ | UNDOV (*Interjection*) | hoorah! (*when you win*) |

Table 1. All the proposed POS Tags for the Uzbek language with their meaning and example words.

The main reason behind the rule-based POS-tagging choice in this work is because of the lack of labelled data big enough to feed the neural network models to expect a good accuracy results. In fact, the output of this tagger can be used as a source for modern POS neural network models.

Apart from that, when POS tagging is applied to languages with rich morphology and agglutination, the rule-based approach is more effective in tagging unfamiliar words (Anbananthen et al., 2017).

---
[1] Universal Dependencies POS-tags and guidelines:
https://universaldependencies.org/u/pos/

**Uzbek language.** Uzbek is a Turkic language spoken by over 40 million people, primarily in Uzbekistan as an official and in neighboring countries as a second language.

The official script of the language is Latin, but the old Cyrillic script is still in use both in official and unofficial basis. The language is, like other Turkic languages in the same family, highly-agglutinative, with SOV word order and does not poses neither gender nor articles. It has been influenced by both the Persian and Russian languages due to historical and cultural interactions[2].

Despite its significant number of speakers, Uzbek is considered a low-resource language in the field of NLP. This is because there is a limited amount of labeled data and resources available for Uzbek language, making it difficult to develop and evaluate NLP models for this language.

The POS-tagger tool created in this research work was assessed using the new dataset, and the experiment results show that the tool has achieved at least 85% accuracy in every field, reaching almost 90% average accuracy for the overall dataset.

The lack of resources for the Uzbek language makes it a challenging task for NLP researchers, however, it also presents an opportunity to contribute to the field by developing NLP models for Uzbek and other low-resource languages. This research aims to provide a valuable resource for NLP tasks in Uzbek, as the presented dataset and the tagging tool, to our best knowledge, is the first of its kind for the low-resource Uzbek language.

## 2. Related Work

POS tagging has evolved over the years, starting with rule-based systems that relied on hand-written grammar rules to identify the POS of words (Voutilainen, 2003). These systems were limited in their accuracy and were not able to handle the complexity and variability of natural language.

With the advent of machine learning, statistical models were developed to automatically learn the POS tags from annotated corpora. These models, such as Hidden Markov Models (HMM) (Kupiec, 1992) and Conditional Random Fields (CRF) (Awasthi et al., 2006; Constant et al., 2011), have improved the accuracy of POS tagging.

With the recent advancements in deep learning, neural network-based models have been developed that have further improved the accuracy and efficiency of POS tagging, using both word-level (Meftah et al., 2018; Perez-Ortiz et al., 2001) and character level representations (dos Santos et al., 2014) of POS tagging.

Related work in the field of NLP for the Uzbek language has primarily focused on the development of resources such as WordNet (K. A. Madatov et al., 2022), datasets for sentiment analysis (Kuriyozov et al., 2022; Matlatipov et al., 2022), as well as semantic evaluation (Salaev et al., 2022b). However, there has been a rapid growth on the development of NLP tools for Uzbek, such as stopword removal (K. Madatov et al., 2022), transliterator(U Salaev et al., 2022), and stemmer (Sharipov & Salaev, 2022; Sharipov & Yuldashov, 2022) recently.

A specific work about Uzbek POS-tagging by Abjalova and Iskandarov (Abdurashetona et al., 2021) also propose 12 tags for the Uzbek parts of speech. Currently, the field of NLP is developing rapidly and playing an important role in solving problems in scientific, economic and cultural fields (Sharipov, Mattiev, et al., 2022).

## 3. Methodology

This section is devoted to the methodological part of the research work, starting from the details of the tagging algorithm used, followed by the text normalization steps, all the way till the corpus creation and the annotation process.

### 3.1 Tagging algorithm

A specific algorithm was used to properly create the POS-tagger tool: Given Uzbek text is first tokenized into sentence and then word levels, then tokens(words) are searched from the dictionary of lemmas[3] from an existing previous work (Sharipov & Sobirov, 2022) and other available sources like Apertium package for Uzbek[4]. If the lemma is found, the word class corresponding to it is determined accordingly. In the case of a token (word) being found in a dictionary in more than one class, then the sequence of suffixes of this token (word) are taken and searched in the dictionary of suffixes.

If there are no suffixes in the word and there is a problem in determining which word group it belongs to, in that case, our proposed algorithm determines the category of the current word depending on the category of the words surrounding it (words that are coming after and before it). Let's have a look at the following example: "*Yaxshi ovqat yesang, yaxshi ishlaysan.*" (If you eat well, you work well). The first word *"yaxshi"* (good) in this sentence is an adjective, the second word *"yaxshi"*(well) is an adverb here. There are no suffixes at the word *"yaxshi"*, therefore, it is not possible to determine which category this word belongs to based on its suffixes, so we determine the class of the word using the neighboring words. If the word has more than one possible tag, then rule-based taggers use hand-written rules to identify the correct tag.

A dictionary of more than 80,000 Uzbek words was created alongside their 12 POS tags in an XML format were created as the main source of the UzbekTagger tool. Besides, special rules were developed to identify two words with the same tag. Some of the created rules the tagger contains are listed below as an example:
- IF previous word's POS is adjective THEN the current POS is noun;
- IF previous word's POS is adverb THEN current POS is verb;
- IF next WORD takes yordamchi fel THEN current [with next] one gets the verb POS;

---

[2] More about the Uzbek language:
https://en.wikipedia.org/wiki/Uzbek_language
[3] As the definition of lemma varies among the existing NLP research works, the term in this paper is refer as the dictionary form of word.

[4] Apertium monolingual package for Uzbek:
https://github.com/apertium/apertium-uzb

- IF previous WORD_SUF is egalik THEN current POS is noun;
- IF current WORD_SUF is verb_suffix THEN current POS is verb;
- IF current WORD_SUF is noun_suffix THEN current POS is noun;
- IF current WORD is bog`lovchi THEN previous and next POS is the same;

### 3.2. Normalization

During the creation of the tagger tool, we encountered several problems with text normalization. There are 29 letters and 1 apostrophe (') in the Uzbek language's official Latin script. Two of them are these letters: **o'** and **g'**, in texts, there are cases where these two letters' sign are replaced by a apostrophe: o' and g', or completely different characters are used: o', o`, g', g`. In such cases, tokenizers tokenize incorrectly. Let's tokenize this sentence: "O`qituvchi gapirdi", tokens: "O", "qituvchi", "gapirdi", but in the correct form this sentence must consist of two tokens: "O'qituvchi", "gapirdi". In Uzbek, the apostrophe does not come after the letters o and g. Therefore, in solving this problem, we changed all the signs after o and g to (') [*similar to the number 6*], and in other cases to (') [*similar to the number 9*]. Below are some examples:

o`rdak->o'rdak,  (duck)
g'ildirak->g'ildirak,  (wheel)
ta`lim->ta'lim,  (education)

| № | Category | Sentences | Words |
|---|---|---|---|
| 1 | Adabiyot (Literature) | 76 | 999 |
| 2 | Anatomiya (Anatomy) | 60 | 1020 |
| 3 | Biologiya (Biology) | 87 | 1001 |
| 4 | Botanika (Botany) | 59 | 1014 |
| 5 | Din tarixi (History of relegion) | 67 | 1016 |
| 6 | Dunyo (World) | 74 | 1006 |
| 7 | Fizika (Physics) | 81 | 1008 |
| 8 | Geografiya (Geography) | 61 | 1002 |
| 9 | Huquq (Law) | 57 | 1014 |
| 10 | Informatika (Informatics) | 84 | 1005 |
| 11 | Iqtisodiyot (Economy) | 38 | 1027 |
| 12 | Jamiyat (Society) | 44 | 1003 |
| 13 | Kimyo (Chemstry) | 75 | 1002 |
| 14 | Madaniyat (Culture) | 72 | 1000 |
| 15 | Matematika (Mathematics) | 43 | 999 |
| 16 | Ona tili (Mother tongue) | 98 | 1012 |
| 17 | Qishloq xo'jaligi (Agriculture) | 69 | 1006 |
| 18 | Siyosat (Politics) | 54 | 1305 |
| 19 | Sport (Sports) | 78 | 1008 |
| 20 | Tarix (Hitory) | 85 | 1005 |
| 21 | Texnologiya (Technology) | 74 | 1005 |
| 22 | Tibbiyot (Medicine) | 52 | 1013 |
| 23 | Zoologiya (Zoology) | 93 | 1012 |
| | **Total:** | **1581** | **23482** |

Table 2. Number of sentences and words per category in the created corpus.

### 3.3. Corpus annotation

One of the most important factors that show the true performance of a POS Tagger is the corpus it was used to asses. In particular, the size of the corpus, the reliability of the tagged corpus, and the diversity of the corpus have a great effect (Can et al., 2021).

Due to the lack of openly-available Uzbek corpus that is diverse enough and is equally balanced over the different fields, we developed a tagged corpus that is evenly distributed across different categories. The raw text was obtained from books openly available at the Republican Youth E-Library[5] and the category the text belongs to was assigned based on the field the book belongs to.

The tagged dataset contains 23 categories with total number of 1581 sentences made of 23482 words in total. An average of 1000 words were taken from almost all fields available. The detailed composition of the categories, and number of sentences taken are presented in Table 2.

The same POS-tags were used and the same guidelines as the POS-tagger tool were followed during the annotation process. Four annotators with an expert-level linguistic knowledge of Uzbek annotated the created corpus over the course of six months. Each sentence was assured to be annotated at least by two individuals to overcome the human error. The problem of sentences with conflicting tags was solved by a group discussion to choose the right tags.

The choice of so many POS tags for the annotation was a result of an effort to cover all possible word forms as much as possible. This way, the annotated text will avoid possible misconceptions among homonyms. For instance, in the field of biology, the word "tut" (*mulberry*) has to be a noun in the sense of a fruit, and in the field of sports, the word "tut" (*catch*) has to be a verb in the sense of an action.

## 4. Experimental results

For the experiments, we checked the performance of the created POS-tagger tool using the annotated dataset as a source of evaluation. The UzbekTagger tool, which was made as a Python library was fed with raw sentences taken from the annotated corpus, then the output from the tagger was compared with the manually annotated format of the same sentence.

As an additional mean of evaluation the authors also considered the category the sentence belongs to, so that the overall analysis allows to identify on which categories there is a need for more work. Accuracy was chosen as the main metric of evaluation.

To explain the comparison of the tagger output and the manual annotation, let us take an example sentence from the category of Informatics:

*"Mantiqiy formulalar rostlik jadvallari yordamida izohlanadi." (Logical formulas are interpreted using truth tables.)*

The output of the UzbekTagger is as follows:

*"Mantiqiy/NOUN formulalar/NOUN rostlik/NOUN jadvallari/NOUN yordamida/NOUN izohlanadi/VERB ./PUNCT"*

In this example, the first word "Mantiqiy" [*Logical*] is not actually a NOUN, rather it should be an ADJ, so this case was counted as one mistake. The only condition is that if the same word appears wrongly tagged more than once, it was still considered as one mistake.

---
[5] The Republican Youth E-Library: https://kitob.uz

The total mistakes were then calculated, and the detailed performance results over each category are reported in Table 3.

| № | Category | Mistakes | Accuracy |
|---|---|---|---|
| 1 | Adabiyot (Literature) | 126 | 87.40% |
| 2 | Anatomiya (Anatomy) | 31 | 96.97% |
| 3 | Biologiya (Biology) | 96 | 90.41% |
| 4 | Botanika (Botany) | 32 | 96.85% |
| 5 | Din tarixi (History of relegion) | 79 | 92.23% |
| 6 | Dunyo (World) | 56 | 94.44% |
| 7 | Fizika (Physics) | 86 | 91.47% |
| 8 | Geografiya (Geography) | 178 | 82.24% |
| 9 | Huquq (Law) | 157 | 84.52% |
| 10 | Informatika (Informatics) | 115 | 88.56% |
| 11 | Iqtisodiyot (Economy) | 96 | 90.66% |
| 12 | Jamiyat (Society) | 92 | 90.83% |
| 13 | Kimyo (Chemstry) | 100 | 90.00% |
| 14 | Madaniyat (Culture) | 88 | 91.20% |
| 15 | Matematika (Mathematics) | 82 | 91.80% |
| 16 | Ona tili (Mother tongue) | 168 | 83.40% |
| 17 | Qishloq xo'jaligi (Agriculture) | 205 | 79.50% |
| 18 | Siyosat (Politics) | 120 | 90.81% |
| 19 | Sport (Sports) | 101 | 89.99% |
| 20 | Tarix (Hitory) | 72 | 92.84% |
| 21 | Texnologiya (Technology) | 118 | 88.26% |
| 22 | Tibbiyot (Medicine) | 144 | 85.79% |
| 23 | Zoologiya (Zoology) | 58 | 94.27% |
|   | **Total:** | **2400** | **89.78%** |

Table 3. Number of mistakes and accuracy in each category

The results show that the POS-tagger tool performs with at least 83% accuracy in all categories of Uzbek text, with up to 97% accuracy in some others. This indicates that the tagger tool has already includes terminology from fields like Anatomy, Botany and Math, while the terminology has to be enriched for some other fields like Agricultur, Mother tounge, and Law.

## 5. Discussion

When solving the problem of tagging Uzbek language texts by word classes, there is a difficulty in determining the POS tag when the same words appear in sentences in different word classes. For example, in the sentence "*U juda qattiq ishlar edi*" [He/She was very hard-working], the word "*ishlar*" (work) is a verb, but in the sentence "*Kechagi bo'lib o'tgan ishlar yaxshi emas*" (Things happened yesterday were not good), the word "*ishlar*"(things) is a noun. When tagging words in the above case, it is necessary to make a conclusion by knowing which word class the word that comes before and after tagging a word belongs to.

According to the results of the experts' analysis in Table 3, Agriculture showed the lowest Accuracy of 79.50%, while Botany showed the highest Accuracy of 96.85%. The main reason for this kind of a trend happening in the performance can be explained by the scope of the terminology words included in the tagger's stems dictionary, which has to be improved, especially for the fields the tagger is struggling with.

## 6. Conclusion and future work

In conclusion, we presented the first publicly available POS-tagged dataset with more than 1500 sentences, annotated using a balanced Uzbek corpus. Also, the first openly available rule-based Uzbek POS-tagger tool was introduced alongside the dataset that achieved high accuracy results when tested on the annotated dataset. The UzbekTagger tool achieved about 90% overall accuracy over the dataset with more than 20 fields.

In the future, the researchers plan to improve the performance of the POS tagger by incorporating machine learning and neural network techniques. Additionally, the researchers aim to expand the annotated dataset to include more data from different fields. Furthermore, the researchers plan to develop more sophisticated NLP tools for Uzbek, such as dependency parser using the POS-annotated dataset and the tagger tool which will provide more comprehensive NLP support for the Uzbek language.

### Data availability.

The developed a Python-based POS-tagger tool for the Uzbek language is available to be installed and used via the Python Package Index (PyPi)[6]. Apart from that, both the source code of the POS-tagger tool and the annotated dataset files can be found at the project GitHub repository[7]. To our best knowledge, no open-source POS-tagger tool has been created or made publicly available.

### Acknowledgements.

This research work was fully funded by the REP-25112021/113 - "UzUDT: Universal Dependencies Treebank and parser for natural language processing on the Uzbek language" subproject funded by The World Bank project "Modernizing Uzbekistan national innovation system" under the Ministry of Innovative Development of Uzbekistan.

### Declarations.

The authors declare no conflict of interest. The founding sponsors had no role in the design of the study; in the collection, analyses, or interpretation of data; in the writing of the manuscript, and in the decision to publish the results.

---

[6] https://pypi.org/project/UzbekTagger.

[7] https://github.com/MaksudSharipov/UzbekTokenizer.